\def\year{2021}\relax
\documentclass[letterpaper]{article} 
\usepackage{aaai21}  
\usepackage{times}  
\usepackage{helvet} 
\usepackage{courier}  
\usepackage[hyphens]{url}  
\usepackage{graphicx} 
\usepackage{natbib}  
\usepackage{caption} 
\usepackage[colorlinks=true]{hyperref} 
\usepackage{enumerate}
\usepackage{bm}
\usepackage{amsfonts}
\usepackage{amsmath, amssymb}
\usepackage{multirow}
\usepackage{tabularx}
\usepackage{mathrsfs}
\usepackage{amsthm}
\usepackage{booktabs}

\newcommand{\ie}{\emph{i.e.},~}
\newcommand{\eg}{\emph{e.g.}~}
\newcommand{\wrt}{\emph{w.r.t.}~}

\newcommand{\xiaok}[1]{\left(#1\right)}
\newcommand{\zhongk}[1]{\left[#1\right]}
\newcommand{\dak}[1]{\left\{#1\right\}}
\newcommand{\jiaok}[1]{\left<#1\right>}
\newcommand{\shuk}[1]{\left\lVert#1\right\rVert}

\newcommand{\argmax}[1]{{\mathop{\arg\mathrm{max}}_{#1}\,}}

\newcommand{\T}{\top}

\newcommand{\biset}[1]{\{0,1\}^{#1}}

\newcommand{\tr}{\mathrm{tr}}

\newcommand{\diag}{\mathrm{diag}}
\newcommand{\cov}{\mathrm{cov}}

\newcommand{\opseq}[3]{{#1_1 #3 #1_2 #3 \cdots #3 #1_{#2}}}
\newcommand{\seq}[2]{\opseq{#1}{#2}{,}}

\newcommand{\bma}{\bm{a}}
\newcommand{\bmb}{\bm{b}}
\newcommand{\bmc}{\bm{c}}

\newcommand{\bmq}{\bm{q}}
\newcommand{\bmr}{\bm{r}}
\newcommand{\bms}{\bm{s}}
\newcommand{\bmt}{\bm{t}}

\newcommand{\bmv}{\bm{v}}

\newcommand{\bmx}{\bm{x}}

\newcommand{\bmzero}{\bm{0}}

\newcommand{\bmepsilon}{\bm{\epsilon}}

\newcommand{\bmpi}{\bm{\pi}}

\newcommand{\bmA}{\bm{A}}
\newcommand{\bmB}{\bm{B}}
\newcommand{\bmC}{\bm{C}}

\newcommand{\bmR}{\bm{R}}
\newcommand{\bmS}{\bm{S}}
\newcommand{\bmT}{\bm{T}}

\newcommand{\bmW}{\bm{W}}

\newcommand{\bmSigma}{\bm{\Sigma}}

\newcommand{\calG}{\mathcal{G}}

\newcommand{\calL}{\mathcal{L}}

\newcommand{\calN}{\mathcal{N}}
\newcommand{\calO}{\mathcal{O}}

\newcommand{\calQ}{\mathcal{Q}}
\newcommand{\calR}{\mathcal{R}}
\newcommand{\calS}{\mathcal{S}}
\newcommand{\calT}{\mathcal{T}}

\newcommand{\calW}{\mathcal{W}}

\newcommand{\calY}{\mathcal{Y}}

\newcommand{\bbR}{\mathbb{R}}

\def \WSDHQ {WSDHQ}

\makeatletter
\newcommand{\printfnsymbol}[1]{%
  \textsuperscript{\@fnsymbol{#1}}%
}
\makeatother

\title{Weakly Supervised Deep Hyperspherical Quantization for Image Retrieval}
\author{
    Jinpeng Wang\textsuperscript{\rm 1,\rm 2}\thanks{Equal contribution.},
    Bin Chen\textsuperscript{\rm 1}\printfnsymbol{1}\thanks{Corresponding authors.}, Qiang Zhang\textsuperscript{\rm 3}, Zaiqiao Meng\textsuperscript{\rm 4}, Shangsong Liang\textsuperscript{\rm 2}\printfnsymbol{2}, Shutao Xia\textsuperscript{\rm 1}
    \\
}
\affiliations{
    \textsuperscript{\rm 1}Tsinghua Shenzhen International Graduate School, Tsinghua University\\
    \textsuperscript{\rm 2}School of Computer Science and Engineering, Sun Yat-sen University\\
    \textsuperscript{\rm 3}University College London \\
    \textsuperscript{\rm 4}University of Cambridge\\

    wjp20@mails.tsinghua.edu.cn,\ \  cb17@tsinghua.org.cn,\ \ 
    qiang.zhang.16@ucl.ac.uk,\\ 
    zm324@cam.ac.uk,\ \ 
    liangshangsong@gmail.com,\ \ 
    xiast@sz.tsinghua.edu.cn

}

\begin{document}

\maketitle

\begin{abstract}
Deep quantization methods have shown high efficiency on large-scale image retrieval. 
However, current models heavily rely on ground-truth information, hindering the application of quantization in label-hungry scenarios. A more realistic demand is to learn from inexhaustible uploaded images that are associated with informal tags provided by amateur users. 
Though such sketchy tags do not obviously reveal the labels, they actually contain useful semantic information for supervising deep quantization. To this end, we propose \textbf{W}eakly-\textbf{S}upervised \textbf{D}eep \textbf{H}yperspherical \textbf{Q}uantization (\textbf{\WSDHQ{}}), which is the first work to learn deep quantization from weakly tagged images. Specifically, 
\textbf{1}) we use word embeddings to represent the tags and enhance their semantic information based on a tag correlation graph. 
\textbf{2}) To better preserve semantic information in quantization codes and reduce quantization error, we jointly learn semantics-preserving embeddings and supervised quantizer on hypersphere by employing a well-designed fusion layer and tailor-made loss functions. 
Extensive experiments show that \WSDHQ{} can achieve state-of-art performance on weakly-supervised compact coding.
Code is available at \url{https://github.com/gimpong/AAAI21-WSDHQ}.
\end{abstract}

\noindent With the explosive growth of media data on the web, many retrieval tasks need to handle large-scale and high-dimensional data. Due to high computational efficiency and low memory overhead, \emph{learning to hash}~\cite{2018L2H}, as a technique in Approximate Nearest Neighbor (ANN) search, has been applied in many applications. 
Briefly, the goal of hashing is to transform high-dimensional data into compact binary codes while preserving semantic information. 
Based on the ways of measuring distance between encoded data, hashing methods can be roughly categorized into two types. 
\textbf{1}) \emph{Binary hashing}~\cite{gionis1999similarity,salakhutdinov2009semantic} transforms high-dimensional data into hash codes in Hamming space such that distances are computed with fast bitwise operators. 
\textbf{2}) \emph{Quantization}~\cite{jegou2010product,ge2013optimized,babenko2014additive,zhang2014composite,yang2020mean} divides high-dimensional data space into disjoint cells and approximately represents each point by its cell centroid. Since the pairwise distances between data points are pre-computed by inter-centroid distances and stored in a lookup table, the search speed is accelerated.

Recently, deep learning has been integrated into hash models and yields superior performance over the shallow methods~\cite{zhao2015deep,Huang_2019_ICCV,jin2020ssah}. Empirically, deep quantization methods have more powerful representation capability than deep binary hashing methods for ANN search~\cite{cao2016deep,cao2017deep,liu2018deep,chen2019similarity,yuan2020central}. 
Nevertheless, there are two concerns about existing deep quantization methods.
\textbf{1}) \emph{Data hunger}: existing deep quantization methods heavily rely on high-quality supervision. However, despite the advent of large-scale annotated datasets such as ImageNet~\cite{Deng2009ImageNet}, the lack of well-labeled data in specific domains remains a critical bottleneck of deep learning. Collecting massive data with exact labels is usually labor-intensive and expensive, which hinders the deployment of deep quantization methods in practical large-scale applications, \eg search engines and social media. 
\textbf{2}) \emph{High norm variance of deep features}: deep networks often produce representation vectors with relatively high variance on the norm, which adversely degrades the quality of quantization~\cite{wu2017multiscale,eghbali2019deep}.

\begin{figure}[!t]
  \centering
  \includegraphics[width=\linewidth]{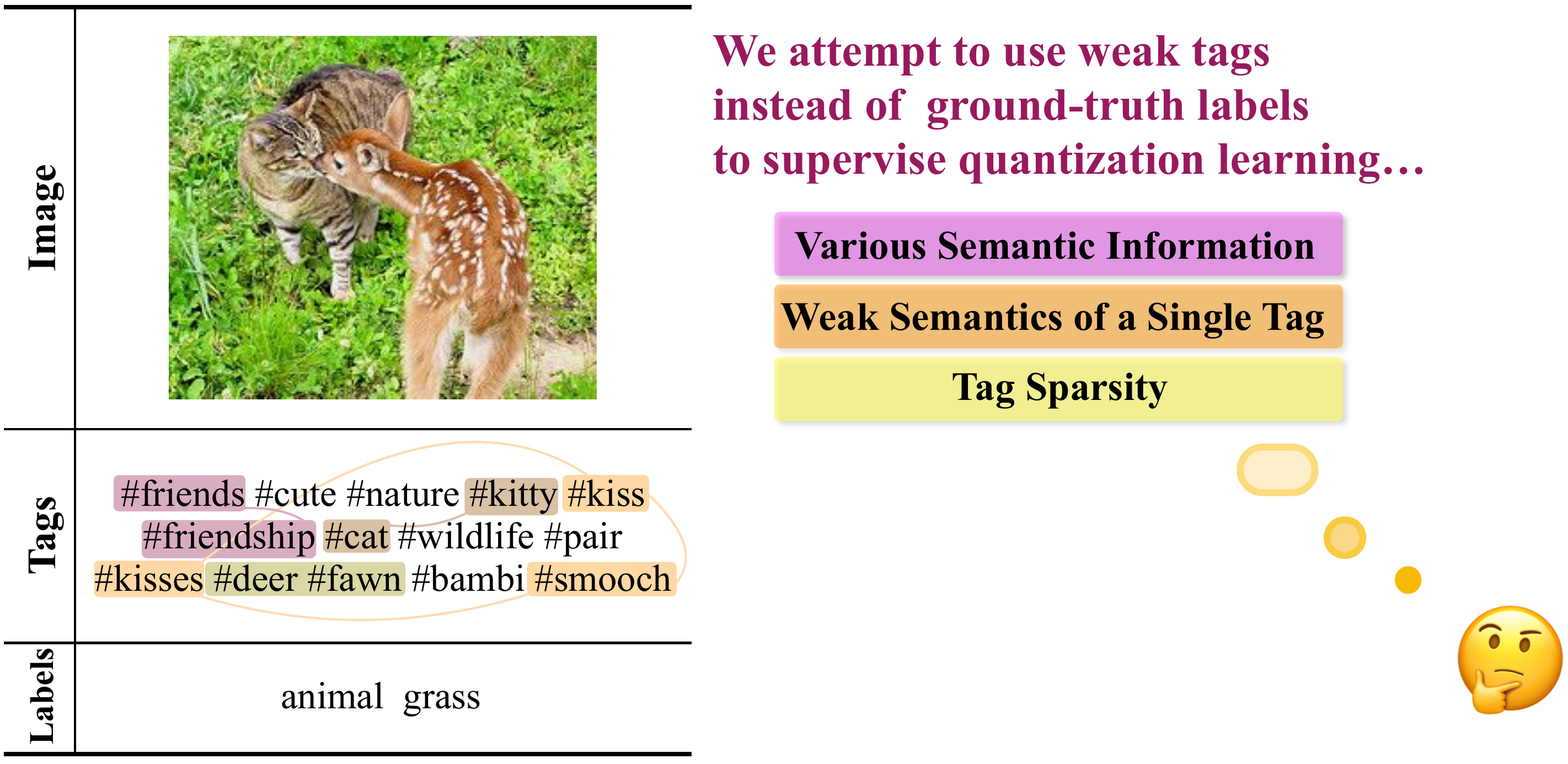}
  \caption{An example from NUS-WIDE dataset to illustrate the problem of weakly supervised quantization using tags.}
  \label{chanllanges}
\end{figure}

To overcome the heavy dependence of manually annotated data, we consider taking the freely available web images with impure tags as training data, and study the novel problem of weakly supervised deep quantization, as illustrated in Fig.~\ref{chanllanges}.
Different from existing deep supervised quantization methods that leverage clear labels as supervision, we try to tackle the following challenges: 
\textbf{1}) \emph{Tag sparsity}: the tag words are nearly unrestricted, so many synonymous tags share similar meanings; the total number of tags could be huge. 
\textbf{2}) \emph{Weak semantics of a single tag}: a single tag can be semantically vague, leading to confusion in learning. 
\textbf{3}) \emph{Various semantics in one image}: an image may contain multiple concepts.
Besides, to further improve deep quantization, we also attempt to reduce the high norm variance of deep embeddings in the quantization model.

In this paper, we propose \textbf{W}eakly-\textbf{S}upervised \textbf{D}eep \textbf{H}yperspherical \textbf{Q}uantization (\textbf{\WSDHQ{}}) for learning to quantize with weak tags. 
To our best knowledge, \WSDHQ{} is the \emph{first} work to address the problem of weakly-supervised deep quantization without using ground-truth labels. It explores the possibility of disconnecting the deep quantization evolution from the scaling of human-annotated datasets, given free and inexhaustible web and social media data.
We make the following contributions in \WSDHQ{}: 
\begin{enumerate}[1)]
    \item On the specific task of image quantization, we are the first to consider enhancing the weak supervision of tags.
    Concretely, we build a tag embedding correlation graph, to effectively enhance tag semantics and reduce sparsity.
    
    \item To reduce the error of deep quantization, we remove the norm variance of the deep features by applying $\ell_2$ normalization and maps visual representations onto a semantic hypersphere spanned by tag embeddings.
    
    \item We further improve the ability of quantization model to better preserve semantic information into quantization codes by designing a novel adaptive cosine margin loss and a novel supervised cosine quantization loss, that directs the training of our model in an end-to-end manner.
    
    \item Extensive experiments show that \WSDHQ{} yields state-of-art retrieval results in weakly-supervised scenario.
\end{enumerate}

\section{Related Work}
\label{sec:relatedWork}

\noindent\textbf{Deep Quantization.}
Deep quantization methods cooperating with CNNs~\cite{krizhevsky2012imagenet,he2016deep} have shown superior performance against traditional non-deep methods that use hand-crafted features~\cite{jegou2010product,ge2013optimized,kalantidis2014locally,babenko2014additive,zhang2014composite,martinez2016revisiting,martinez2018lsq}. 
The goal of deep supervised quantization is to learn compact codes through deep networks that are faithful to given semantic information such as pointwise~\cite{cao2017deep,eghbali2019deep,yuan2020central}, pairwise~\cite{cao2016deep,chen2019similarity} or triplet labels~\cite{yu2018product,liu2018deep}. 
Despite the promising performance, existing methods largely rely on high-quality labels to learn satisfactory models. It limits the application of deep quantization on many real-world scenarios, where a lot of data is available without adequate ground-truths. Different from previous works, we attempt to solve the novel problem of weakly-supervised deep quantization in this paper.

\noindent\textbf{Tackle Norm Variance for Quantization.}
It has been revealed that many adopted deep networks produce representations with relatively large norm variance, which leads to greater quantization error and unexpected performance degeneration~\cite{wu2017multiscale,eghbali2019deep}.
To reduce the norm variance, MSQ~\cite{wu2017multiscale} quantized the data norms using an additional scalar quantizer before applying product quantization (PQ). Nevertheless, it is hard to balance the budget between two quantizers, and PQ is inferior to other quantization methods due to its orthogonality assumption on codebooks~\cite{babenko2014additive}.
DSQ~\cite{eghbali2019deep} removed the norm variance for deep representations by quantizing them on a unit-norm sphere, and learned to close the Euclidean distance between each image embedding and its unique class center.

In this paper, we employ a transformation layer with $\ell_2$ normalization as DSQ did, to embed deep image representations onto a semantic hypersphere spanned by tag embeddings. Different from DSQ, we learn quantization in a multi-semantics (\ie multi-label and multi-class) settings by jointly optimizing two novel cosine losses, \ie adaptive cosine margin loss and supervised cosine quantization loss.

\noindent\textbf{Weakly-supervised Hashing.}
There has been less attention paid to weakly-supervised hashing, where meta-data (weak tags) attached to web images is freely available to be an inexhaustible source of weak supervision~\cite{gomez2018learning}.
To our knowledge, there has been no weakly-supervised quantization method for ANN search until now.
Although there have been a few initial attempts~\cite{zhang2016discrete,Tang2018WMH,guan2018tag,gattupalli2019weakly,cui2020efficient} in weakly-supervised binary hashing, they either rely on pure labels to rectify the model or directly train the model using raw tags without effective semantic enhancement, which leaves a lot of room for improvement. 
\citet{zhang2016discrete} used collaborative filtering to predict tag-label associations, where labels were requested. 
\citet{guan2018tag} proposed a two-stage framework consisting of weakly supervised pre-training and fine-tuning using ground-truth labels. 
Technically, the above two methods are not ideal to be \emph{truly} weakly-supervised hashing, because of the dependence on ground-truth information. 
Weakly Supervised Multimodal Hashing (WMH)~\cite{Tang2018WMH} is the first holistically weakly-supervised hashing method, which constructs binary matrix of tags and formulates hashing as an eigenvalue problem. It tends to be vulnerable because WMH  directly uses weak tags as binary labels.
Weakly Supervised Deep Hashing using Tag Embeddings (WDHT)~\cite{gattupalli2019weakly} employs Word2Vec~\cite{mikolov2013efficient} embeddings for tags. Although noises and vagueness are partially alleviated in WDHT, the tags of each image are roughly represented as the average vector of tag embeddings by simple weighting strategies, which is too rough to grasp various semantics.


Our \WSDHQ{} is the \emph{first} weakly supervised deep quantization method. Different from existing weakly supervised binary hashing methods, there are two improvements: 
\textbf{1}) \WSDHQ{} investigates the semantic relations of weak tags, based on which, it further enhances tag semantics and reduces similar tags confusion so as to improve the supervision.
\textbf{2}) \WSDHQ{} preserves more fine-grained semantic representations during training, which helps to get better performance. We will discuss the effects of these two strategies (\ie loss designs) in the ablation experiment.

\begin{figure*}[!t]
  \centering
  \includegraphics[width=\textwidth]{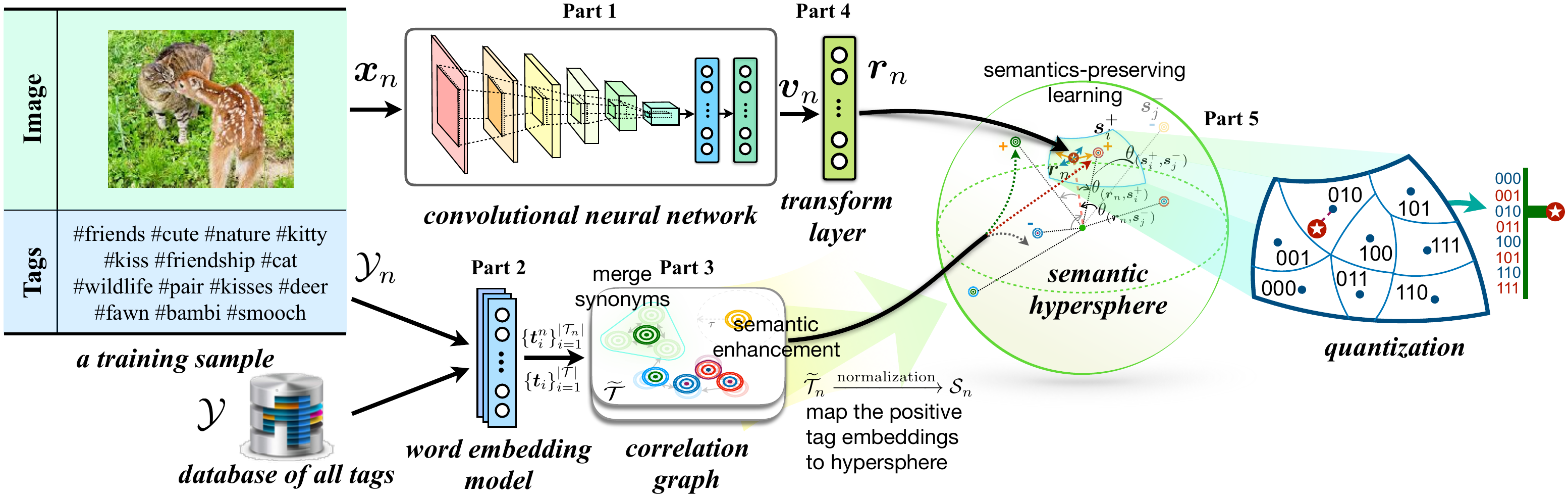}
   \caption{The proposed \textbf{W}eakly \textbf{S}upervised \textbf{D}eep  \textbf{H}yperspherical \textbf{Q}uantization (\textbf{\WSDHQ{}}) consists of five main parts: 1) a standard CNN, 2) a word embedding model, 3) a correlation graph, 4) a transform layer and 5) a semantic hypersphere.}
  \label{method_arc}
\end{figure*}

\section{Proposed Approach}
\label{sec:proposedApproach}
In this section, we first formulate the research problem and give a brief overview of our approach. Then we present the supervision enhancement and the quantization on hypersphere. Finally, we introduce the overall learning algorithm.

\subsection{Problem Formulation}
In a weakly-supervised image retrieval scenario, we are given a training set of $N$ images with attached tag sets $\dak{\bmx_n,\calY_n}_{n=1}^N$, where each image is represented as a $P$-dimensional vector $\bmx_n\in\bbR^P$ and is associated with a set of textual tags $\calY_n\subset\calY$, $\calY$ is a set containing all tags. While the web images with tags are easy to obtain from search engines or social media, the raw tags are not reliable enough to be ground-truth labels.
Thus the goal of weakly-supervised deep quantization is to learn a compositional quantizer $q:\bmx\in\bbR^P\mapsto\bmb\in\biset{B}$, which encodes each point $\bmx$ into a compact $B$-bit binary code $\bmb$ by preserving the weak semantics from tags via DNNs.

\subsection{Overview of Our Approach}
This paper enables efficient image retrieval by presenting \textbf{W}eakly \textbf{S}upervised \textbf{D}eep \textbf{H}yperspherical \textbf{Q}uantization (\textbf{\WSDHQ{}}) in an end-to-end deep learning architecture, as shown in Fig.~\ref{method_arc}.
\WSDHQ{} consists of five main components:
\textbf{1}) A standard CNN $f$, \eg AlexNet~\cite{krizhevsky2012imagenet}, to extract deep features of images.
\textbf{2}) A word embedding model, \eg Word2Vec~\cite{mikolov2013efficient}, to represent each tag $y_i\in \calY$ as an embedding $\bmt_i$, while $\calT$ and $\calT_n$ are tag embedding sets \wrt raw tag sets $\calY$ and $\calY_n$.
\textbf{3}) A correlation graph $\calG$, to enhance the semantics of tags and merge synonymous tags. We denote semantically enhanced tag sets via $\calG$ as $\widetilde{\calT}$ and $\widetilde{\calT}_n$.
\textbf{4}) A transform layer $g$, to embed deep features $\bmv_n$ as $\bmr_n$ on a hypersphere, which is spanned by the normalized tags $\calS$.
\textbf{5}) A semantic hypersphere, on which the norm variances of data points are reduced to zeros. With the help of two well-customized cosine losses, our joint process of semantics-preserving learning for image embedding $\bmr_n$ and quantization on the hypersphere yield better compact codes for ANN search.

\subsection{Enhance Weak Supervision with Tag Correlation}
\label{subsec:corrGraph}
When using attached tags as supervision, we first consider the issues of \emph{weak semantics of single tags} and \emph{tag sparsity}.

\noindent\textbf{Semantic Correlation Graph.}
We first extract the pre-trained Word2Vec embedding $\bmt_i\in\bbR^D$ for each informative tag $y_i\in\calY$, where $D$ is the dimension of textual embedding space, then form the tag embedding sets $\calT$ and $\dak{\calT_n}_{n=1}^N$ \wrt $\calY$ and $\dak{\calY_n}_{n=1}^N$. Let $\bmT\in\bbR^{D\times|\calT|}$ be the tag embedding matrix. Next we obtain a $k$-nearest neighbor set $\text{NN}^k(i)$ by cosine similarity for each tag $\bmt_i$. A semantic correlation graph $\calG$ is then constructed on the total tag set $\calT$ by an adjacency matrix $\bmA=(a_{ij})\in\biset{|\calT|\times|\calT|}$, where
\begin{gather*}
  a_{ij}=\begin{cases}
    1, & \text{if $j\in\text{NN}^k(i)$ and $\frac{\bmt_i^\T\bmt_j}{\shuk{\bmt_i}\shuk{\bmt_j}}\ge\tau$, or just $i=j$,}\\
    0, & \text{otherwise,}
  \end{cases}
\end{gather*}
and $\tau$ is the correlation threshold that determines whether a neighbor tag is related to an anchor tag in semantics. Note that each row $\bma_i$ in $\bmA$ indicates the semantic similarities between $\bmt_i$ and other tags on $\calG$.

\noindent\textbf{Semantic Enhancement on Graph.}
Since the neighbor information from $\calG$ can effectively enhance the semantic representations of single tags and alleviate their semantic biases, we can obtain an semantics-enhanced tag embedding matrix $\widetilde{\bmT} = \bmT\tilde{\bmA}^\T$ by aggregating the embeddings from its neighbor tags, 
where $\tilde{\bmA}$ is the row-wise normalization of $\bmA$.

\noindent\textbf{Reduce Sparse Tags.}
After semantic enhancement, the representations of similar tags are expected to move closer and aggregate in small regions, so those sparse and redundant tags can be further detected and merged. 
The merging process follows classic density-based clustering algorithm~\cite{DBSCAN96}. 
Any $\bmt_i',\bmt_j'$ on $\calG$ are \emph{density-reachable} to each other if $d(\bmt_i',\bmt_j')<\epsilon$, where $\epsilon$ is the merging threshold and $d(\cdot,\cdot)$ is a distance metric, \eg the $\ell_2$ distance. 
Each time we pick up one unprocessed point $\bmt_i'$ from $\calG$ in order and compute its density-reachable set $\calR(i,\epsilon)=\{\bmt_j'\mid \bmt_j'$ is density-reachable from $\bmt_i'\}$. 
Once $|\calR(i,\epsilon)|>1$, we merge and replace all the elements in $\calR(i,\epsilon)$ by the average point.
Finally, we obtain refreshed embedding sets  $\widetilde{\calT}$ and $\{\widetilde{\calT}_n\}_{n=1}^N$ by removing same embeddings in each tag set.

\subsection{Quantization on Semantic Hypersphere}
\label{subsec:quantization}
We adopt a standard CNN $f:\bbR^P\mapsto\bbR^V$ for extracting $V$-dimensional deep features $\bmv_n=f(\bmx_n)$ of image $\bmx_n$. 
$\calS = \dak{{\bmt_i'}/{\|\bmt_i'\|_2}}_{i=1}^{|\widetilde{\calT}|}$ and $\calS_n=\dak{{\bmt_i'}/{\|\bmt_i'\|_2}}_{i=1}^{|\widetilde{\calT}_n|}$ are the normalized total tag set and the normalized tag set of image $\bmx_n$, which span the $D$-dimensional hyperspherical semantic space.
We set up a transform layer $g:\bbR^V\mapsto\bbR^D$ with $\ell_2$ normalization to remove norm variance of deep features $\bmv_n$ and map $\bmv_n$ onto the semantic hypersphere as $\bmr_n=g(\bmv_n)=\frac{\sigma(\bmW_g\bmv_n)}{\|\sigma(\bmW_g\bmv_n)\|_2}$, where $\sigma(\cdot)$ is the activation, \eg the hyperbolic tangent (tanh), and $\bmW_g$ is the parameter matrix of $g$. 
For brevity, we use $h=f\circ g$ to denote the fusion of network $f$ and layer $g$ such that $\bmr_n=h(\bmx_n)$.

\noindent\textbf{Semantics-preserving Learning on Hypersphere.}
We propose an adaptive cosine margin loss $\calL_n$ to enable semantics-preserving learning on hypersphere, namely
\begin{equation}\nonumber
\calL_n=\sum_{\bms^+_i\in\calS_n}\sum_{\bms^-_j\in\calN_n}\zhongk{\Delta_{ij}-\cos\theta_{(\bms^+_i,\bmr_n)}+\cos\theta_{(\bms^-_j,\bmr_n)}}_{+},
\end{equation}
where $\zhongk{\cdot}_+=\max(0,\cdot)$,  $\cos\theta_{(\bmv,\bmv')}=\langle\bmv,\bmv'\rangle,\,\text{s.t. }\|\bmv\|_2=\|\bmv'\|_2=1$,
$\calN_n$ is the negative semantic set of $\bmx_n$ and
$\Delta$ is the cosine margin. 
This metric loss encourages the image embedding $\bmr_n$ to move closer to positive semantic embeddings while pushing it from negative embeddings.

Ideally, we would like to utilize all negative semantic embeddings of $\bmx_n$, \ie $\calN_n=\calS\backslash\calS_n$ for learning, but this can lead to unacceptably high computational cost as $|\calS|$ is usually huge in real world. In practice, we only involve $K_n$ most tricky negative semantic embeddings as our negative set, \ie \begin{equation}\nonumber
    \calN_n=\argmax{\bms^-_i\in\calS\backslash\calS_n,\,|\calN_n|=K_n}{\cos\theta_{(\bms^-_i,\bmr_n)}}.
\end{equation}

To keep the image representations consistent with precise semantic information, we adopt an adaptive margin strategy $\Delta_{ij}$ that depends on the discrepancy between positive and negative semantics. For example, 
a smaller margin $\Delta_\text{(cat, dog)}$ is acceptable while $\Delta_\text{(kiss, dog)}$ should adapt to be larger, since the semantic representation of negative semantics ``\emph{dog}'' is closer to that of positive semantics ``\emph{cat}'' than positive semantics ``\emph{kiss}''.
Specifically, we define
\begin{equation*}
\Delta_{ij}=2^{1-\gamma}\cdot\xiaok{1-\cos\theta_{(\bms_i^+,\bms_j^-)}}^\gamma,
\end{equation*}
where $\gamma$ is a hyper-parameter such that smaller $\gamma$ leads to larger adaptive margin under same semantic similarity.

\noindent\textbf{Supervised Cosine Quantization.} We propose a supervised quantizer on hypersphere to enable efficient image retrieval.
Specifically, each image embedding $\bmr_n$ will be quantized with a set of $M$ codebooks $\bmC=\zhongk{\seq{\bmC}{M}}$, each $\bmC_m=\{\bmc_{m1},\bmc_{m2},\cdots,\bmc_{mK}\}$ contains $K$ codewords, and each codeword $\bmc_{mk}$ is a $D$-dimensional centroid vector obtained by $k$-means. 
The codeword assignment vector $\bmb_n$ is segmented into $M$ $1$-of-$K$ indicator vectors $\bmb_n=\zhongk{\bmb_{1n};\bmb_{2n};\cdots;\bmb_{Mn}}$ \wrt $M$ codebooks, and each one-hot indicator vector $\bmb_{mn}$ indicates which one of $K$ codewords in the $m$-th codebook is used to compose the approximation for $\bmr_n$. 
The proposed quantizer encodes each image embedding $\bmr_n$ as the sum of $M$ codewords, each of which comes from its codebook $C_m$ assigned by an indicator vector $\bmb_n$, \ie $\bmr_n\approx\hat{\bmr}_n\equiv\sum_{m=1}^M\bmC_m\bmb_{mn}$. 
Then we integrate semantic supervision into quantization learning, with the cosine quantization loss formulated as
\begin{equation}\nonumber
  \calQ_n=\sum_{\bms_i\in\calS}\xiaok{\cos\theta_{(\bms_i,\bmr_n)}-\cos\theta_{(\bms_i,\hat{\bmr}_n)}}^2.
\end{equation}

\noindent\textbf{Approximate Nearest Neighbor Search.} Approximate nearest neighbor (ANN) search by maximum inner-product similarity (MIPS) is a powerful tool for quantization methods~\cite{cao2017deep,liu2018deep}. 
Note that on the unit hypersphere, the cosine similarity between two points can be equivalently transformed into inner-product. 
Hence, with the image database of $N$ quantized binary codes $\{\bmb_n\}_{n=1}^N$, we adopt \emph{Asymmetric Quantizer Distance} (AQD)~\cite{jegou2010product} as the metric, which computes the cosine of the angle on hypersphere between a given query $\bmq$ and the reconstruction of a database point $\bmx_n$ as
\begin{equation*}
\text{AQD}(\bmq,\bmx_n)=\cos\theta_{(\bmr_q,\hat{\bmr}_n)}=\bmr_q^\T\xiaok{\sum_{m=1}^M\bmC_m\bmb_{mn}},
\end{equation*}
where $\bmr_q$ is the hyperspherical embedding \wrt query $\bmq$. 
We set up a query-specific lookup table of $M\times K$ items for $\bmq$, which stores the pre-computed results of inner-product between $\bmq$ and all codewords in $\bmC$. Hence, the AQD can be efficiently computed by summing chosen items from lookup table according to the quantization code $\bmb_n$.

\subsection{Learning Algorithm}
\noindent\textbf{Overall Objective.} \WSDHQ{} enables efficient image retrieval in an end-to-end architecture, which jointly learns semantics-preserving hyperspherical embedding and supervised quantization in a total objective as
\begin{equation}\label{objective}
\min_{\calW,\bmC,\bmB}\sum_{n=1}^N\xiaok{\calL_n+\lambda \calQ_n},
\end{equation}
where $\lambda$ is a positive  hyper-parameter to balance the adaptive cosine margin loss $\calL$ and the semantically supervised quantization loss
$\calQ$, and $\calW$ denotes the learnable network parameters. Through optimizing objective~(\ref{objective}), \WSDHQ{} preserves the semantic information of tags into hyperspherical embeddings while effectively reducing quantization error. 

There are three sets of variables in objective~(\ref{objective}): \textbf{1}) the network parameters $\calW$, \textbf{2}) $N$ binary codes $\bmB=\zhongk{\seq{\bmb}{N}}=\zhongk{\opseq{\bmB}{M}{;}}$, where $\bmB_m=\zhongk{\bmb_{m1},\bmb_{m2},\cdots,\bmb_{mN}}$ is the subcode matrix of all points \wrt the $m$-th codebook $\bmC_m$ and \textbf{3}) $M$ codebooks $\bmC=\zhongk{\seq{\bmC}{M}}$. We adopt an commonly used alternating optimization paradigm~\cite{liu2018deep}, which iteratively optimizes one variable set while fixing the others.\\
\noindent\textbf{Learning $\calW$.} Many back-propagation algorithms can be adopted to optimize the network parameters $\calW$. With the advance of automatic differentiation techniques in mainstream machine learning libraries~\cite{abadi2016tensorflow,paszke2019pytorch}, it is easy to optimize $\calW$ within a few code lines.

\noindent\textbf{Learning $\bmB$.}
We update $N$ quantization codes $\bmB$ by fixing $\calW$ and $\bmC$ as known variables. Since each $\bmb_n$ is independent with $\{\bmb_{n'}\}_{n'\ne n}$, the optimization of $\bmB$ can be split into $N$ subproblems, e.g., we optimize $\bmb_n$ as
\begin{gather}
\label{updateB}
\min_{\bmb_n}\xiaok{\bmr_n-\sum_{m=1}^M\bmC_m\bmb_{mn}}^\T\bmSigma_{\bmS}\xiaok{\bmr_n-\sum_{m=1}^M\bmC_m\bmb_{mn}},\\
\nonumber
\text{s.t. }\|\bmb_{mn}\|_0=1,\,\bmb_{mn}\in\biset{K},
\end{gather}
where $\bmSigma_{\bmS}=\sum_{\bms_i\in\calS}\bms_i{\bms_i}^\T$ is the covariance matrix of semantic embeddings, which reflects the latent distribution of queries as all visual features will be eventually embedded to the hypersphere spanned by these embeddings.
Objective~(\ref{updateB}) is generally an NP-hard problem of discrete combinatorial optimization, while some stochastic local search algorithms can provide a promising solution. We take the widely used  Iterated Conditional Modes (ICM) algorithm~\cite{besag1986statistical,zhang2014composite} to solve it.
 Given fixed $\{\bmb_{m'n}\}_{m'\ne m}$, we update $\bmb_{mn}$ by exhaustively checking all codewords in $\bmC_m$ and finding the codeword such that objective~(\ref{updateB}) is minimized. 
In each encoding iteration, the $M$ indicators $\{\bmb_{mn}\}_{m=1}^M$ are calculated alternatively in this way. Moreover, we inject some stochastic relaxations~\cite{zeger1992globally,martinez2018lsq} into ICM to avoid local optimum so as to improve the performance of quantization. 
Specifically, for the $i$-th encoding iteration, we add an iteration-decaying perturbation $\bmpi(i)=(T(i)/M)\cdot\bmepsilon$ to codebooks $\bmC$ and get perturbed codebooks $\tilde{\bmC}=\bmC+\bmpi(i)$, where $T(i)=\sqrt{1-(i/I)}$ is the temperature scheduled for the $i$-th iteration among all $I$ encoding iterations, $\bmepsilon\sim\calN(\bmzero,\bmSigma)$ and $\bmSigma=\diag(\cov(\bmR))$ is the diagonal covariance proportional to $\bmR$. 
Finally, we replace the codebook $\bmC_m$ in objective~(\ref{updateB}) with $\tilde{\bmC}_m$ before learning $\{\bmb_{mn}\}_{n=1}^N$.

\noindent\textbf{Learning $\bmC$.} We update $M$ codebooks $\bmC$ by fixing $\calW$ and $\bmB$ as known variables, and rewrite the objective~(\ref{objective}) as
\begin{equation}
\label{learnC}
\min_{\bmC}\tr\xiaok{(\bmR-\bmC\bmB)^\T\bmSigma_{\bmS}(\bmR-\bmC\bmB)},
\end{equation}
where $\bmR=\zhongk{\seq{\bmr}{N}}$ is the image embedding matrix. 
The optimal solution of objective~(\ref{learnC}) comes in a closed form as $\bmC=\bmR\bmB^\T(\bmB\bmB^\T)^{-1}$. 
Note that the binary matrix $\bmB$ satisfies structurally regular conditions: 
\textbf{1}) Diagonal $\bmB_i^\T\bmB_i$ can be computed as histograms of the codes in $\bmB_i$, and $\bmB_i^\T\bmB_j=(\bmB_j^\T\bmB_i)^\T$ can be computed as bivariate histograms of the codes in $\bmB_i$ and $\bmB_j$.
\textbf{2}) $\bmR\bmB_i$ can be computed by treating the columns of $\bmB_i$ as binary vectors that select the columns of $\bmR$ to sum together. 
With the help of these properties, the optimal solution of $\bmC$ can be solved with time complexity $\calO(MND)$ and $\calO(M^2N)$, compared with their na\"ive computations of $\calO(MKND)$ and $\calO(M^2K^2N)$ \cite{martinez2018lsq}. 

\section{Experiments}

\label{experiments}
In this section, we conduct extensive experiments to evaluate our proposed \WSDHQ{} model with several state-of-art shallow and deep hashing methods on two web image datasets.

\subsection{Setup}
To our best knowledge, there are only two large-scale and commonly-used web image datasets (\textbf{MIR-FLICKR25K}, \textbf{NUS-WIDE}) that contain image-tag-label triplets. (\emph{E.g.}, \textbf{ImageNet} does not contain tag information.) Thus we conduct our empirical evaluations on them.

\textbf{MIR-FLICKR25K}~\cite{huiskes08} is a dataset of 25,000 Flickr images associated with 1,386 tags. The authors labelled the images with 38 semantic concepts in total, which are only used for evaluation in our experiments. 2,000 images are randomly sampled as test queries and the rest are used as retrieval database and training images.

\textbf{NUS-WIDE}~\cite{nus-wide-civr09} is a large-scale web image dataset also collected from Flickr, which contains 269,648 images with 5,018 tags provided by users. The authors have manually annotated each image with a pre-defined set of 81 ground-truth labels, which are only used for evaluation. We collect a subset of 193,752 images with the 21 most frequent labels for experiments. We follow~\cite{cao2017deep,liu2018deep} to randomly sample 5,000 images as queries and remain the rest as the database, from which we further sample 10,000 images and their tag sets as training data.

Following standard evaluation protocols adopted in previous works~\cite{cao2017deep,liu2018deep,gattupalli2019weakly}, we use three evaluation metrics: Mean Average Precision (\textbf{MAP}), Precision-Recall curves (\textbf{PR}), and Precision curves \wrt the number of top returned results (\textbf{P@N}). To make fair comparisons, all methods use identical training and test sets, and we follow previous works to adopt MAP@5000 for both datasets. Given a query, the ground truth is defined as: if a result shares at least one common label with the query, it is relevant; otherwise it is irrelevant.

We compare the retrieval performance of the proposed \textbf{\WSDHQ{}} model with several state-of-art hashing methods, including: \textbf{1}) Five shallow unsupervised methods, \textbf{LSH}~\cite{gionis1999similarity}, \textbf{SH}~\cite{weiss2009spectral}, \textbf{SpH}~\cite{Lee2012SpH}, \textbf{ITQ}~\cite{6296665} and \textbf{AQ}~\cite{babenko2014additive}. \textbf{2}) One deep unsupervised method, \textbf{DeepBit}~\cite{Lin_2016_CVPR}. \textbf{3}) Two shallow weakly-supervised methods, \textbf{WMH}~\cite{Tang2018WMH} and \textbf{WDH}~\cite{cui2020efficient}. \textbf{4}) One deep weakly-supervised method, \textbf{WDHT}~\cite{gattupalli2019weakly}.

We use AlexNet~\cite{krizhevsky2012imagenet} to extract 4096-dimensional deep $fc7$ features from each image for shallow models. For deep models, we directly use raw image pixels as input and adopt AlexNet ($conv1\sim fc7$, pre-trained on ImageNet) as the backbone network. We take the Word2Vec~\cite{mikolov2013efficient} as word embedding model and represent each tag with a 300-dimensional pre-trained embedding.

We implement \WSDHQ{} based on TensorFlow~\cite{abadi2016tensorflow}. For the semantic correlation graph, we set the maximum number  of neighbors $k=20$ for each tag, the correlation threshold $\tau$ as $0.75$ and the merging threshold $\epsilon$ as $0.1$. We set the number of tags for negative tags selected in the adaptive cosine margin loss as $K_n=1000$. We fine-tune all layers copied from pre-trained model and train the transform layer via back-propagation from scratch. We adopt a mini-batch Adam with default parameters as optimizer. Besides, we select learning rate from $10^{-5}\sim10^{-2}$, the hyper-parameter $\lambda$ from $10^{-5}\sim10^{-1}$ and $\gamma$ from $[0.3,0.5,0.7,1,2,3,4]$ via cross-validation. Following~\cite{cao2016deep,cao2017deep,liu2018deep,eghbali2019deep}, we adopt $K=256$ codewords for each codebook, thus the binary index for each image of all $M$ codebooks requires $B=M\log_2K=8M$ bits (\ie $M$ bytes).

\begin{table}[t]
\centering
\resizebox{\columnwidth}{!}{
\begin{tabular}{l*{7}{c@{~}}c}
    \toprule\multirow{2}{*}{Dataset} & \multicolumn{4}{c}{\phantom{00}MIR-FLICKR25K} & \multicolumn{4}{c}{NUS-WIDE} \\
    \cmidrule(l){2-5} \cmidrule(l){6-9}
     & 8 bits\phantom{0} & 16 bits\phantom{0} & 24 bits\phantom{0} & 32 bits\phantom{0} & 8 bits\phantom{0} & 16 bits\phantom{0} & 24 bits\phantom{0} & 32 bits\phantom{0} \\
    \midrule
    LSH\phantom{0} & 0.524\phantom{0} & 0.570\phantom{0} & 0.562\phantom{0} & 0.572\phantom{0} & 0.376\phantom{0} & 0.392\phantom{0} & 0.413\phantom{0} & 0.418\phantom{0} \\
    SH\phantom{0} & 0.592\phantom{0} & 0.609\phantom{0} & 0.617\phantom{0} & 0.604\phantom{0} & 0.498\phantom{0} & 0.505\phantom{0} & 0.477\phantom{0} & 0.492\phantom{0} \\
    SpH\phantom{0} & 0.556\phantom{0} & 0.582\phantom{0} & 0.579\phantom{0} & 0.586\phantom{0} & 0.463\phantom{0} & 0.448\phantom{0} & 0.464\phantom{0} & 0.461\phantom{0} \\
    ITQ\phantom{0} & 0.641\phantom{0} & 0.623\phantom{0} & 0.654\phantom{0} & 0.633\phantom{0} & 0.536\phantom{0} & 0.545\phantom{0} & 0.556\phantom{0} & 0.563\phantom{0} \\
    AQ\phantom{0} & 0.637\phantom{0} & 0.645\phantom{0} & 0.658\phantom{0} & 0.661\phantom{0} & 0.524\phantom{0} & 0.567\phantom{0} & 0.587\phantom{0} & 0.592\phantom{0} \\
    \midrule
    DeepBit\phantom{0} & 0.628\phantom{0} & 0.632\phantom{0} & 0.623\phantom{0} & 0.608\phantom{0} & 0.542\phantom{0} & 0.555\phantom{0} & 0.558\phantom{0} & 0.552\phantom{0} \\
    \midrule
    WMH\phantom{0} & 0.656\phantom{0} & 0.684\phantom{0} & 0.672\phantom{0} & 0.671\phantom{0} & 0.558\phantom{0} & 0.592\phantom{0} & 0.605\phantom{0} & 0.601\phantom{0} \\
    WDH\phantom{0} & 0.669\phantom{0} & 0.678\phantom{0} & 0.694\phantom{0} & 0.685\phantom{0} & 0.577\phantom{0} & 0.602\phantom{0} & 0.618\phantom{0} & 0.627\phantom{0} \\
    \midrule
    WDHT\phantom{0} & 0.704\phantom{0} & 0.733\phantom{0} & 0.737\phantom{0} & 0.724\phantom{0} & 0.652\phantom{0} & 0.670\phantom{0} & 0.682\phantom{0} & 0.692\phantom{0} \\
    \textbf{\WSDHQ{}} & \textbf{0.744\phantom{0}} & \textbf{0.751\phantom{0}} & \textbf{0.765\phantom{0}} & \textbf{0.772\phantom{0}} & \textbf{0.716\phantom{0}} & \textbf{0.722\phantom{0}} & \textbf{0.738\phantom{0}} & \textbf{0.731\phantom{0}} \\
    \bottomrule
\end{tabular}
}
\caption{Mean Average Precision (MAP) Results for Different Number of Bits on the Two Benchmark Image Datasets.}
\label{tab:overall}
\end{table}

\subsection{Results}
The MAP results of all methods are reported in Table~\ref{tab:overall}, which shows that the proposed \WSDHQ{} model substantially outperforms all the comparison methods. Specifically, compared to AQ (shallow quantization with deep features as input), the best unsupervised hashing method, \WSDHQ{} achieves absolute increases of \textbf{10.8\%}, \textbf{15.9\%} in the average MAP on MIR-FLICKR25K and NUS-WIDE, respectively. Compare to WDHT (deep binary hashing), the state-of-art weakly-supervised hashing method, \WSDHQ{} outperforms WDHT by appreciable margins of \textbf{3.4\%} and \textbf{5.3\%} in average MAP on two datasets, respectively.

We discover several interesting insights from the MAP results. \textbf{1}) The weak tags can actually be utilized as supervision. Weakly-supervised methods (\eg \WSDHQ{} and WDHT) significantly outperform unsupervised methods (\eg AQ and DeepBit). \textbf{2}) The quantization often shows superiority over binary hashing under the same code length, for instance, the shallow unsupervised quantization method AQ achieves better performance than shallow unsupervised binary hashing competitors SpH, SH as well as LSH, and our \WSDHQ{} also outperforms deep weakly-supervised binary hashing WDHT. \textbf{3}) Deep quantization methods (\eg \WSDHQ{}) can learn better codes by jointly preserving semantics and reducing the quantization error, significantly outperforming the shallow counterparts ITQ and AQ with deep features.

\begin{figure}[!t]
  \centering
   \includegraphics[width=\linewidth]{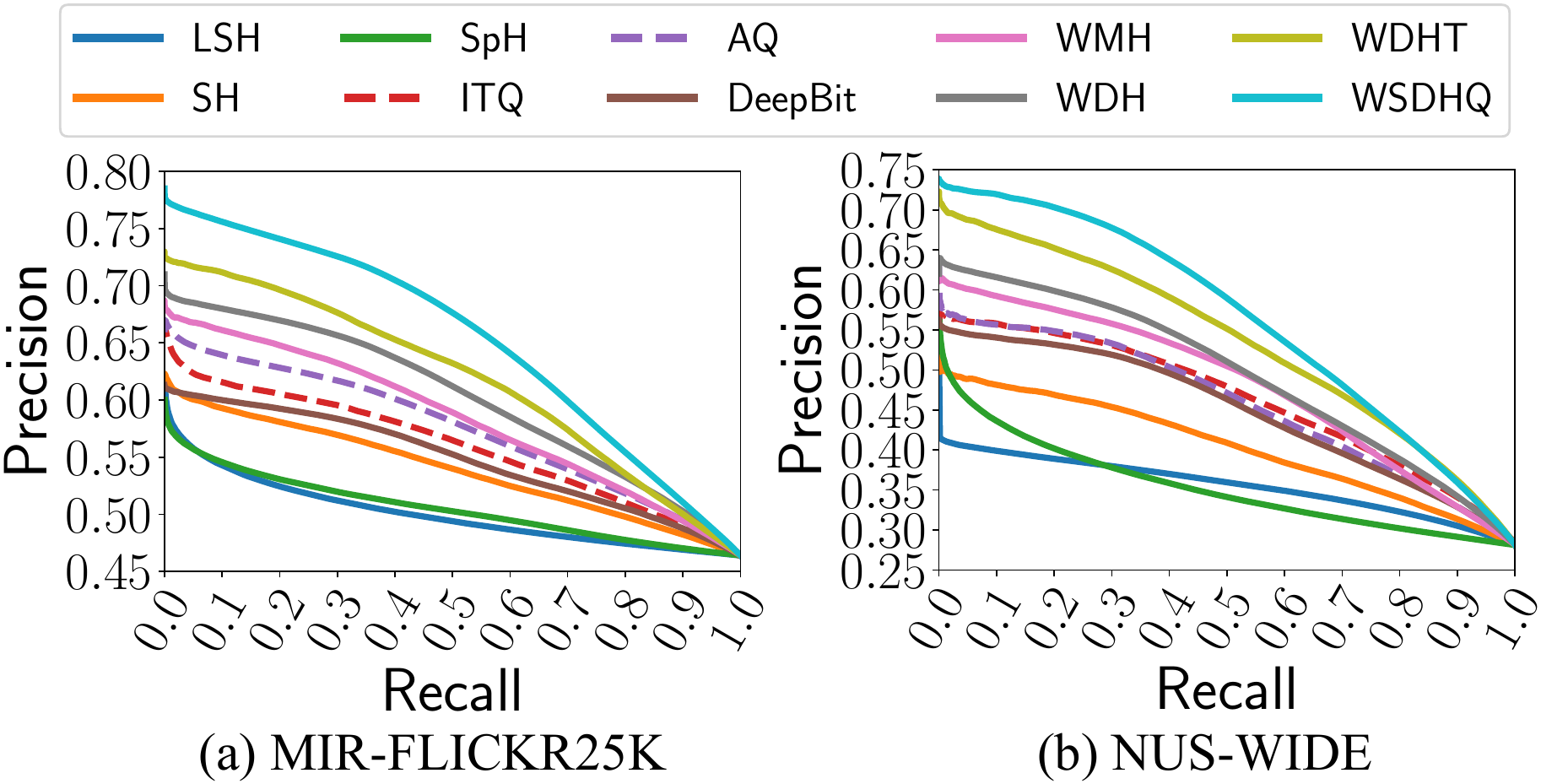}
   \caption{Precision-recall curves on the MIR-FLICKR25K and NUS-WIDE datasets with binary codes @ 32 bits.}
  \label{PR}
\end{figure}

\begin{figure}[!t]
  \centering
   \includegraphics[width=\linewidth]{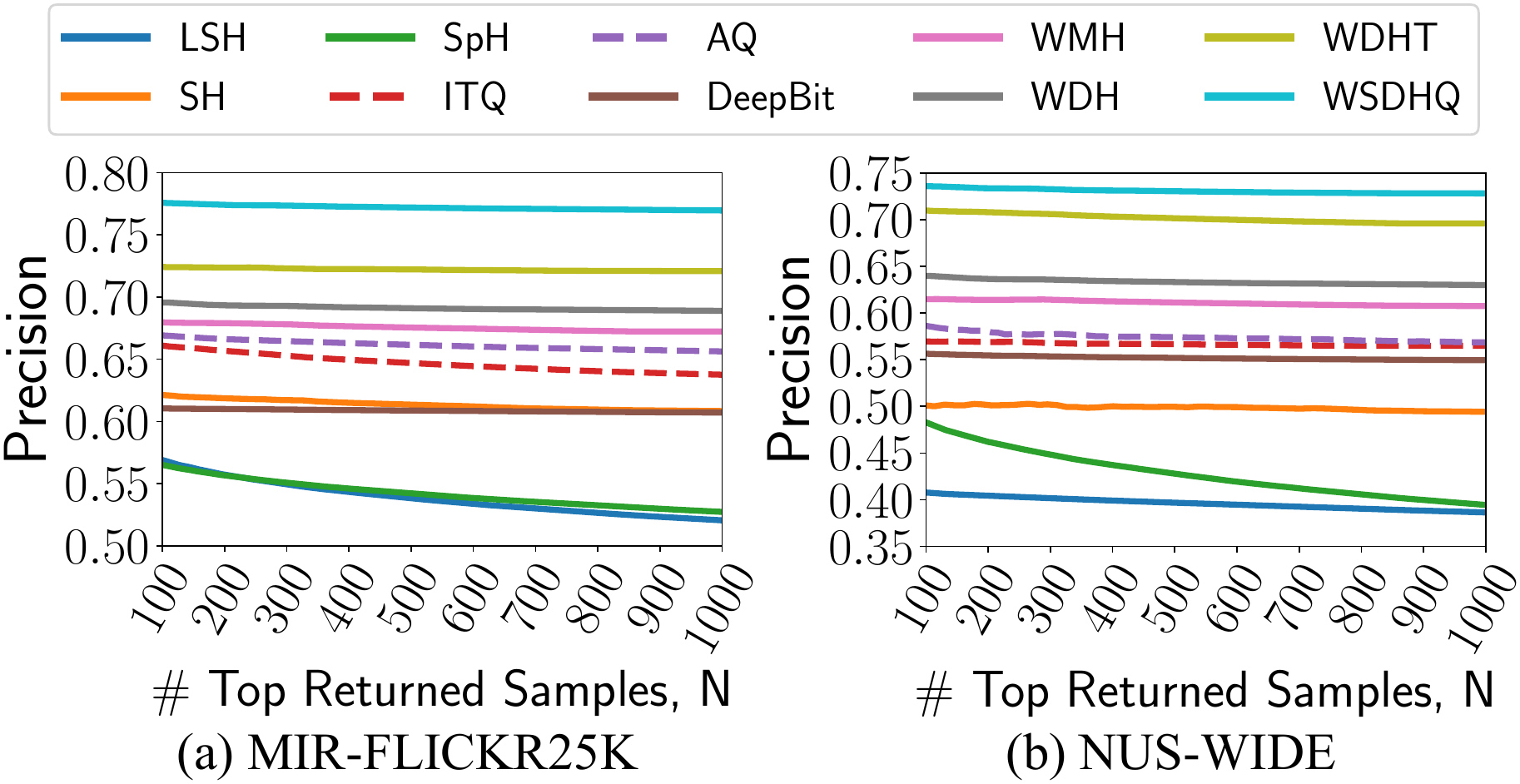}
   \caption{Precision@top-N curves on the MIR-FLICKR25K and NUS-WIDE datasets with binary codes @ 32 bits.}
  \label{P@N}
\end{figure}

The retrieval performance in terms of Precision-Recall curves and Precision curves \wrt different numbers of top returned samples are shown in Figures~\ref{PR} and~\ref{P@N}, respectively. These metrics are widely used in practical image search systems. The proposed \WSDHQ{} model significantly outperforms all comparison methods by large margins in the two metrics. In particular, \WSDHQ{} achieves much higher precision than all compared baselines at low recall levels or when the number of top returned samples is small. This is desirable for precision-oriented retrieval in practical systems, in which users usually pay more attention to the top-$N$ returned results with a relatively small $N$.

\subsection{Ablation Study}

\begin{table}[t]
\centering
\resizebox{1.0\columnwidth}{!}{
\begin{tabular}{l*{7}{c@{~}}c}
\toprule
\multirow{2}{*}{Dataset} & \multicolumn{4}{c}{\phantom{00}MIR-FLICKR25K} & \multicolumn{4}{c}{NUS-WIDE} \\
\cmidrule(l){2-5} \cmidrule(l){6-9}
 & 8 bits\phantom{0} & 16 bits\phantom{0} & 24 bits\phantom{0} & 32 bits\phantom{0} & 8 bits\phantom{0} & 16 bits\phantom{0} & 24 bits\phantom{0} & 32 bits\phantom{0} \\
\midrule
\WSDHQ{}$_\calG$\phantom{0} & 0.728\phantom{0} & 0.736\phantom{0} & 0.749\phantom{0} & 0.751\phantom{0} & 0.708\phantom{0} & 0.711\phantom{0} & 0.725\phantom{0} & 0.717\phantom{0} \\
\WSDHQ{}$_\text{N}$\phantom{0} & 0.724\phantom{0} & 0.727\phantom{0} & 0.740\phantom{0} & 0.755\phantom{0} & 0.703\phantom{0} & 0.716\phantom{0} & 0.723\phantom{0} & 0.726\phantom{0} \\
\WSDHQ{}$_\calL$\phantom{0} & 0.711\phantom{0} & 0.740\phantom{0} & 0.744\phantom{0} & 0.753\phantom{0} & 0.698\phantom{0} & 0.701\phantom{0} & 0.713\phantom{0} & 0.701\phantom{0} \\
\WSDHQ{}$_2$\phantom{0} & 0.725\phantom{0} & 0.736\phantom{0} & 0.731\phantom{0} & 0.747\phantom{0} & 0.706\phantom{0} & 0.707\phantom{0} & 0.702\phantom{0} & 0.694\phantom{0} \\
\textbf{\WSDHQ{}}\phantom{0} & \textbf{0.744\phantom{0}} & \textbf{0.751\phantom{0}} & \textbf{0.765\phantom{0}} & \textbf{0.772\phantom{0}} & \textbf{0.716\phantom{0}} & \textbf{0.722\phantom{0}} & \textbf{0.738\phantom{0}} & \textbf{0.731\phantom{0}} \\
\bottomrule
\end{tabular}}
\caption{Mean Average Precision (MAP) Results of \WSDHQ{} and Its Variants on Two Benchmark Datasets.}
\label{tab:ablation}
\end{table}

We investigate four variants of \WSDHQ{}: 
\textbf{1}) \textbf{\WSDHQ{}$_\calG$}, a variant of \WSDHQ{} that removes the semantic correlation graph $\calG$ as well as the steps of semantics enhancement and synonyms merging (\ie sparse tags reducing) on $\calG$. 
\textbf{2}) \textbf{\WSDHQ{}$_\text{N}$}, a variant which does not involve $\ell_2$ normalization for both tag embeddings and image embeddings. 
Note that this variant conducts semantics-preserving learning no longer on hypersphere, we replace all the cosine similarity $\cos\theta_{(\bmv,\bmv')}$ by inner-product $\jiaok{\bmv,\bmv'}$ between any two $D$-dimensional vectors $\bmv$ and $\bmv'$ in loss functions $\calL_n$ and $\calQ_n$ for all training samples. 
During retrieval, the AQD between any query $\bmq$ and database image $\bmx_n$ is computed with inner-product, \ie $\text{AQD}(\bmq,\bmx_n)=\bmr_q^\T\hat{\bmr}_n$, where $\bmr_q\in\bbR^D$ is the transformed embedding of $\bmq$ and $\hat{\bmr}_n\in\bbR^D$ is the quantization reconstruction of image $\bmx_n$. 
\textbf{3}) \textbf{\WSDHQ{}$_\calL$}, a variant of \WSDHQ{} which replaces the adaptive cosine margin loss for semantics-preserving learning with the mini-batch-wise hinge loss in WDHT~\cite{gattupalli2019weakly} as
\begin{equation}\nonumber
\calL_n=\sum_{\bar{\bms}_j\in\bar{\calS}^b\backslash\{\bar{\bms}_n\}}\max{}(0,\delta+\bar{\bms}_j^\T\bmr_n-\bar{\bms}_n^\T\bmr_n),
\end{equation}
where $\bar{\bms}_n=\frac{1}{|\calS_n|}\sum_{\bms_i\in\calS_n}\bms_i$ is the average point of all the semantic embeddings in $\calS_n$, $\bar{\calS}^b$ is the average semantic embedding set of the $b$-th mini-batch, and $\delta=0.2$ is the pre-defined margin recommended by the authors of WDHT. 
\textbf{4}) \textbf{\WSDHQ{}$_2$}, the two-stage variant which separately learns semantics-preserving embeddings and quantization codes. The MAP results on two datasets are reported in Table~\ref{tab:ablation}.

\noindent\textbf{Enhancing Semantics on Correlation Graph.} \WSDHQ{} outperforms \WSDHQ{}$_\calG$ by 2.3\% and 1.6\% in the average MAP on two datasets, which indicates that the tag processing on $\calG$ plays an important role in weakly-supervised deep quantization, because redundant and synonymous tags in a tag set may lead to semantic bias and degrade performance.

\noindent\textbf{Learning and Quantizing on Hypersphere.} We find that the \WSDHQ{} with $\ell_2$ normalization for norm variance removal boosts 2.9\% and 1.4\% on the average MAP from \WSDHQ{}$_\text{N}$ on two datasets. It shows that the transformation to hypersphere contributes to learn better image representations and reduce the quantization error.

\noindent\textbf{Loss for Visual Embedding Learning.} Another observation is that \WSDHQ{} outperforms \WSDHQ{}$_\calL$ by 2.8\% and 3.3\% in average MAP on two datasets. \WSDHQ{}$_\calL$ adopts a mini-batch-wise hinge loss in previous state-of-art weakly-supervised (binary) hashing method WDHT. The result shows that this loss is not optimal for weakly-supervised quantization (more generally, weakly-supervised hashing), because tag embedding averaging for each image is merely a coarse-grained estimation of true semantics, which fails to model the true tag distribution in semantic space. 
Moreover, the hinge loss may mistakenly push two semantically similar image embeddings far away from each other, since the pairwise similarity is hard to well measure with weak semantics. By contrast, the adaptive cosine margin loss in \WSDHQ{} is more granular. Instead of direct assertions of semantic inter-relationship for training images, it wisely focuses on the pointwise intra-relationships between images and tags in hyperspherical semantic space, thus better preserving semantic information into image embeddings.

\noindent\textbf{Joint Process of Embedding Learning and Quantization.} By jointly learning semantics-preserving image embeddings as well as controlling the quantization error, \WSDHQ{} outperforms \WSDHQ{}$_2$ by 3.2\% and 3.5\% in average MAP on two datasets. This verifies that end-to-end quantization can improve the quantization quality of deep image embeddings.

\subsection{Parameter Sensitivity}
\begin{figure}[t]
  \centering
   \includegraphics[width=\linewidth]{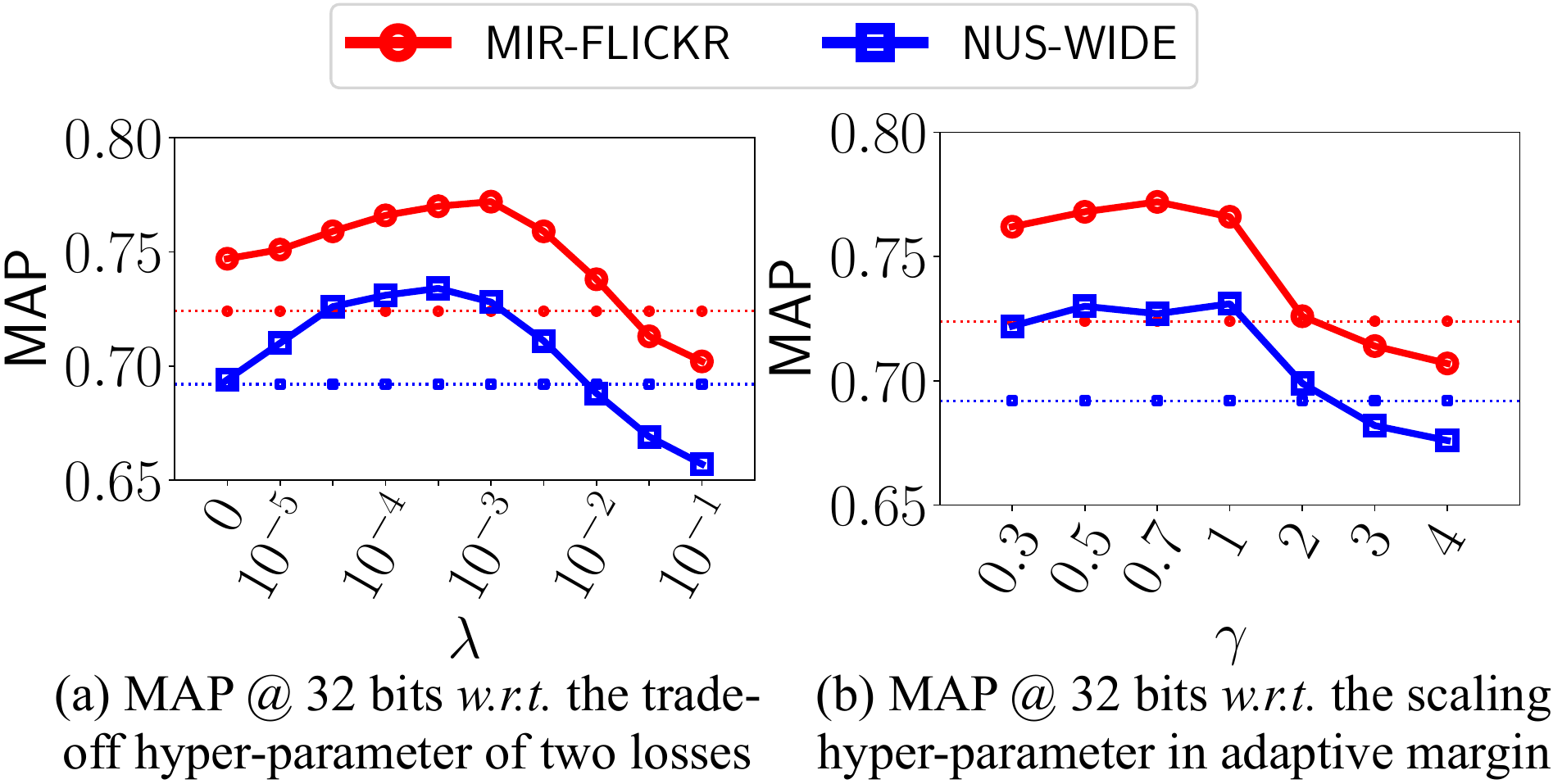}
   \caption{The MAP results of \WSDHQ{} @ 32 bits \wrt $\gamma$ and $\lambda$ on two datasets. The values on dotted lines are the MAP results of WDHT (\ie the best baseline) @ 32 bits.}
  \label{fig:para_sent}
\end{figure}

We study the sensitivities of \textbf{1}) $\lambda$, the trade-off hyper-parameter between losses $\calL_n$ and $\calQ_n$, and \textbf{2}) $\gamma$, the scaling hyper-parameter of the adaptive margin in $\calL_n$. The MAP results are shown in Fig.~\ref{fig:para_sent}. 
\WSDHQ{} surpasses the best baseline within a wide range of $\lambda$ and $\gamma$. The hill-shape trend curves of MAP \wrt $\lambda$ and $\gamma$ confirm the importance of selecting appropriate hyper-parameters in \WSDHQ{}. 
\WSDHQ{} degenerates into its two-stage variant \WSDHQ{}$_2$ that gets inferior results when $\lambda\rightarrow0$, which justifies the effectiveness of joint learning of deep visual embeddings and deep supervised quantization. 
A moderately smaller $\gamma$ that leads to relatively larger margins helps the \WSDHQ{} to learn more discriminative quantization codes, while we should avoid degeneration or even non-convergence caused by too small $\gamma$ with abnormally large margins.

\section{Conclusions}
\label{sec:conclusion}
In this paper we propose the Weakly-Supervised Deep Hyperspherical Quantization (\WSDHQ{}) for efficient image retrieval. 
Different from current deep quantization methods, \WSDHQ{} enables learning quantization codebook from weakly tagged web images without using ground-truth labels. 
The weak supervision is enhanced via semantic enhancement for tags and sparse tag reduction based on the investigated tag correlation. 
Our developed joint learning of deep visual embeddings and semantics-preserving quantizer on hypersphere also yield a significant performance boost to \WSDHQ{} on large-scale image retrieval.
Extensive experiments demonstrate state-of-art retrieval performance on two well-known and widely-tested datasets.
Our work encourages the exploration of weakly-supervised deep quantization by leveraging web and social media data, which promotes quantization to adapt real-world scenario. 
Future work includes improving weakly supervised deep quantization by detecting and completing the probably missing semantic information in the given tag sets during training.

\section{Acknowledgments}
This work is supported in part by the National Key Research and Development Program of China under Grant 2018YFB1800204 and the National Natural Science Foundation of China under Grant 61771273 and Grant 61906219.

\bibliography{main}
\end{document}